\documentclass[runningheads]{llncs}
\usepackage[T1]{fontenc}
\usepackage{amsmath}
\usepackage{amssymb}
\usepackage{multirow}
\usepackage{tabularx}
\usepackage{booktabs}
\usepackage[table]{xcolor}

\newcommand{\modelname}{CAMMST}
\newcommand{\myparagraph}[1]{\noindent\textbf{#1}}

\usepackage{graphicx}
\begin{document}
\title{Contrastive and Adaptive Multi-modal Masked Autoencoder for Spatial Transcriptomics}
\titlerunning{Contrastive and Adaptive Multi-modal MAE for ST}
%
\author{Joohyeok Kim\inst{1} \and
Taejin Jeong\inst{2} \and
Jinyeong Kim\inst{3} \and
Seong Jae Hwang\textsuperscript{\ensuremath{\dagger}}\inst{1}}

\authorrunning{Kim et al.}
%
\institute{
Department of Artificial Intelligence, Yonsei University, Seoul, Republic of Korea \and
Department of Computer Science, Yonsei University, Seoul, Republic of Korea \and
Yonsei University College of Medicine, Seoul, Republic of Korea \\
\email{\{kyyle2114, starforest, jinyeong1324, seongjae\}@yonsei.ac.kr}}
\maketitle              
\begingroup
\renewcommand{\thefootnote}{}
\footnotetext{
\textsuperscript{\ensuremath{\dagger}} Corresponding Author
}
\endgroup

\begin{abstract}
The high cost of spatial transcriptomics (ST) has driven extensive studies into predicting gene expression directly from H\&E histology images.
However, this prediction task faces an inherent limitation, as tissue morphology alone provides insufficient information to fully resolve underlying gene expression.
To address this limitation, a recent study leverages partial gene expression to guide the prediction process alongside histology images.
Building on this paradigm, we approach the prediction task as a spatial imputation problem, employing a Masked Autoencoder (MAE) to utilize a small fraction of gene expression as genetic anchors for inferring whole-slide gene expression profiles.
Specifically, we propose a bio-saliency score and a learning-to-rank strategy to adaptively identify the most informative spots within the tissue. 
Based on these identified spots, our framework selects contiguous regions as genetic anchors to ensure suitability for real-world ST profiling hardware.
To effectively leverage these anchors, we design a cross-modal joint encoder that integrates visual and genetic modalities. 
By aligning the selected anchors with their corresponding visual features via contrastive learning, the encoder generates robust joint representations to accurately predict gene expression across the whole slide.
Notably, our framework consistently surpasses existing methods in both histology-only prediction and spatial imputation, achieving superior accuracy even without genetic anchors and further excelling with as little as 10\% transcriptomic coverage.
Our code is available at \url{https://github.com/Kyyle2114/CAMMST}.


\end{abstract}


\section{Introduction}
Spatial Transcriptomics (ST) has revolutionized tissue analysis by preserving the spatial context of gene expression patterns, thereby enabling the precise profiling of complex biological phenomena~\cite{st}.
Despite this potential, the routine clinical application of ST remains constrained by high costs, the need for domain expertise, and labor-intensive workflows~\cite{st_limitation}.
On the other hand, Hematoxylin and Eosin (H\&E)-stained whole slide images (WSIs) are routinely acquired in clinical pathology, providing an abundant and widely accessible resource.
Accordingly, numerous approaches~\cite{merge,bleep,hist2st,histogene,thitogene,triplex,sepal,feast} have attempted to predict gene expression solely from these WSIs.

However, a recent framework~\cite{translation_completion} indicates that the predictive performance of existing morphology-only methods remains moderate, rendering them far from clinically transferable.
This predictive bottleneck is fundamentally attributed to a lack of sufficient shared information between tissue morphology and gene expression.
Considering biological phenomena such as time uncoupling~\cite{time_uncouple,translation_completion}, where transcriptomic changes precede observable morphological changes, current approaches relying entirely on histological features exhibit clear limitations.

As a practical alternative, a recent study~\cite{ph2st} has proposed a spatial imputation method that utilizes gene expression profiles from a small fraction of tissue spots as anchors to guide the prediction process, demonstrating substantial performance improvements.
Nevertheless, the strategy for selecting these anchors remains to be further refined.
Specifically, relying on non-adaptive sampling strategies, such as random selection, fails to guarantee optimal anchor placement and risks choosing uninformative regions that do not aid prediction.
Furthermore, real-world ST profiling hardware (\textit{e.g.}, $10\times$ Visium, GeoMx) is optimized for capturing contiguous ROIs rather than spatially fragmented points~\cite{hardware}. 
Given these considerations, a practical sampling strategy needs to integrate adaptive, information-driven selection with the requirement of spatial continuity.

To satisfy these prerequisites, a truly practical framework should identify the most informative contiguous regions within the tissue and leverage their measured gene expression as genetic anchors to accurately predict the transcriptomic profiles of the remaining tissue.
Achieving this objective requires an adaptive sampler that identifies the optimal anchor regions, alongside a cross-modal architecture capable of effectively integrating genetic information and morphological context. 
Notably, predicting unmeasured regions from partial measurements naturally aligns with the learning paradigm of Masked Autoencoder (MAE)~\cite{mae}, which infers the complete structure from a visible fraction.
In this context, MAE emerges as an optimal framework for spatial imputation, effectively leveraging measured gene expression and morphological context to achieve accurate transcriptomic prediction.

To this end, we propose \modelname{} (\textbf{C}ontrastive and \textbf{A}daptive \textbf{M}ulti-modal \textbf{M}AE for \textbf{ST}).
Specifically, we introduce a sampler network that adaptively selects contiguous regions to serve as genetic anchors.
We define a bio-saliency score that quantifies local transcriptional deviation, enabling the sampler to identify the most informative regions within the tissue. 
Additionally, we employ a learning-to-rank strategy to train the sampler network, ensuring robustness against cross-slide batch effects by modeling the relative relationships between tissue regions.
Furthermore, we design a cross-modal joint encoder that processes histological images and genetic anchors. 
This encoder aligns measured gene expression with its corresponding histological features via contrastive learning. 
Under the MAE paradigm, these aligned representations are then leveraged to predict whole-slide gene expression profiles.
Experimental results demonstrate that \modelname{} consistently surpasses existing morphology-only baselines and achieves higher accuracy by incorporating as little as 10\% of the total genetic data, offering a practical and high-precision solution for clinical workflows.
\begin{figure}
\includegraphics[width=\textwidth]{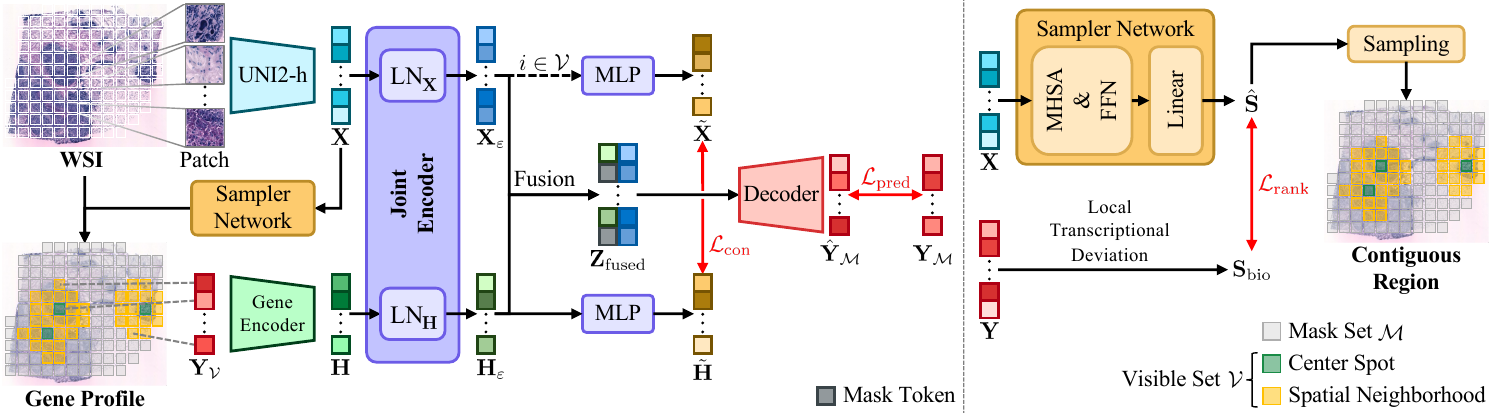}
\caption{\modelname{} adaptively samples contiguous regions as genetic anchors via a bio-saliency guided sampler. A cross-modal joint encoder then integrates visual and genetic modalities via contrastive alignment to predict whole-slide gene expression profiles.}
\label{fig:model}
\end{figure}
\section{Method}
The overall architecture of \modelname{} is illustrated in Fig.~\ref{fig:model}. 
Formally, ST data is represented as a collection of $N$ spots, where each spot $i$ consists of a gene expression vector $\mathbf{y}_i \in \mathbb{R}^{G}$ ($G$ denotes the number of genes) and an image patch $\mathbf{I}_i \in \mathbb{R}^{H \times W \times 3}$. 
These spot-level expression vectors form the complete gene expression set $\mathbf{Y} = \{\mathbf{y}_1, \dots, \mathbf{y}_N\}$, while the corresponding image feature set $\mathbf{X} = \{\mathbf{x}_1, \dots, \mathbf{x}_N\}$ is obtained by extracting a feature vector $\mathbf{x}_i \in \mathbb{R}^{d}$ from each $\mathbf{I}_i$ using UNI2-h~\cite{uni}.
We define the visible set $\mathcal{V}$ as the set of genetic anchors, representing a small subset of spots where gene expression is measured. 
Following the MAE paradigm, the model predicts the gene expression profiles $\hat{\mathbf{Y}}_\mathcal{M} = \{\hat{\mathbf{y}}_i\}_{i \in \mathcal{M}}$ for the masked set $\mathcal{M} = \{1, \dots, N\} \setminus \mathcal{V}$ by leveraging the measured gene expression $\mathbf{Y}_\mathcal{V} = \{\mathbf{y}_i\}_{i \in \mathcal{V}}$ and the morphological context $\mathbf{X}$.

For consistency, we denote the transformer block~\cite{transformer} as $f(\mathbf{Z}; \theta)$, where $\mathbf{Z}$ represents the input sequence and $\theta = \{\theta_\mathrm{attn}, \theta_\mathrm{ffn}\}$ denotes the model parameters.
We employ the Attention with Linear Biases (ALiBi)~\cite{alibi,alibi_2d_titan}, which injects a distance-based bias into the attention operation. This allows the model to encode both local and global contexts based on spot distance, without requiring explicit positional embeddings:
\begin{equation}
   \mathbf{Z}' = \mathbf{Z} + \mathrm{MHSA}_{\mathrm{ALiBi}}(\mathrm{LN}(\mathbf{Z});\theta_\mathrm{attn}), \quad 
   f(\mathbf{Z};\theta) = \mathbf{Z}' + \mathrm{FFN}(\mathrm{LN}(\mathbf{Z}');\theta_\mathrm{ffn}).
\end{equation}

\clearpage
\subsection{Bio-Saliency Region Sampling}~\label{method:bio}

\myparagraph{Bio-Saliency Score.} To identify the most informative regions to serve as genetic anchors, we first require a quantitative metric to evaluate the informativeness across the tissue.
Since our framework utilizes gene expression from only a small fraction of the tissue, these selected regions should encapsulate as much diverse information as possible to guide the prediction.
To this end, we quantify this informativeness based on local transcriptional deviation.
By targeting areas that sharply differ from their spatial surroundings, we ensure the selected regions capture a rich transcriptomic profile.
Notably, this strategy conceptually aligns with SEPAL~\cite{sepal}, which posits that biological variability holds physiological significance.
Accordingly, we define the bio-saliency score for spot $i$ as $S_\mathrm{bio}^{(i)} = D_i T_i$, combining the local deviation $D_i$ with a noise-suppressing gating term $T_i$:
\begin{equation}
	D_i = \left\| \tilde{\mathbf{y}}_i - \frac{1}{|\mathcal{N}(i)|}\sum_{j \in \mathcal{N}(i)} \tilde{\mathbf{y}}_j \right\|_2, 
    \quad
    T_i = \frac{1}{1 + \exp(-0.05 \cdot (C_i - \tau_\mathrm{C}))}.
\end{equation}
Here, $\tilde{\mathbf{y}}_i$ denotes the normalized gene expression vector, and $\mathcal{N}(i)$ represents the set of neighbor spots for spot $i$. 
The gating term $T_i$ utilizes the total expression count $C_i= \sum_{g=1}^G y_{i, g}$ and a threshold $\tau_\mathrm{C}$ to filter out low-signal spots. 
Such spots often exhibit unreliable signals dominated by technical noise; including them as genetic anchors could adversely affect the training process.
To suppress these artifacts, we set $\tau_\mathrm{C}$ to the 20th percentile of the total expression counts.

\myparagraph{Contiguous Region Selection.} 
To identify these highly informative anchors solely from histological images during inference, we design a sampler network~\cite{adamae} trained to predict $S_{\mathrm{bio}}$.
The sampler network estimates importance scores $\hat{\mathbf{s}} \in \mathbb{R}^{N}$ from the morphological context $\mathbf{X}$ to select contiguous regions.
Formally, for a given input $\mathbf{X}$, the sampler network predicts $\hat{\mathbf{s}}$ as:
\begin{equation}
    \mathbf{Z}_\mathrm{samp} = f_{\mathrm{samp}}(\mathbf{X}; \theta_\mathrm{samp}), \quad 
    \hat{\mathbf{s}} = \mathrm{Linear}(\mathbf{Z}_\mathrm{samp}).
\end{equation}
\noindent Using these scores, the sampler identifies $K$ center spots to form regions; $K$ spots are sampled without replacement from a categorical distribution $\mathrm{Cat}(\mathrm{softmax}(\hat{\mathbf{s}}))$ during training, while the top-$K$ spots with the highest scores are selected at inference. 
Each center spot is expanded to include its spatial neighbor spots, forming a contiguous region that adheres to the physical constraints of ST hardware.
The visible set $\mathcal{V}$ is constructed as the union of these selected regions.

\myparagraph{Scale-Aware Ranking Loss.} A significant challenge in training the sampler network is the inherent batch effects across slides; the absolute scale and distribution of $S_{\mathrm{bio}}$ substantially differ among samples due to variations in staining intensity and experimental conditions.
Consequently, directly regressing these absolute scores may bias the model toward slides with dominant intensity ranges or hinder stable convergence.
To address this, we adopt a pairwise ranking strategy inspired by learning-to-rank methodologies~\cite{ranknet}. 
Specifically, we introduce a scale-aware ranking loss that prioritizes the relative ordering of $S_\mathrm{bio}$ between spots over fitting their absolute magnitudes:
\begin{equation}
    \mathcal{L}_\mathrm{rank} = \sum_{(i, j) \in \mathcal{P}} |S_\mathrm{bio}^{(i)} - S_\mathrm{bio}^{(j)}|^\beta \, \log\left(1 + \exp\left(-(\hat{\mathbf{s}}_i - \hat{\mathbf{s}}_j) \, \mathrm{sgn}(S_\mathrm{bio}^{(i)} - S_\mathrm{bio}^{(j)})\right)\right),
\end{equation}
\noindent where $\mathcal{P}$ denotes the set of all possible spot pairs within a single slide, and $\beta$ is a hyperparameter controlling the sensitivity to score differences.

\subsection{Contrastive Cross-Modal Fusion}~\label{method:fusion}
\myparagraph{Cross-Modal MAE.} To effectively integrate the genetic information and morphological context, we design a multi-stream joint encoder building upon CAV-MAE~\cite{cavmae}.
First, for the visible set $\mathcal{V}$ identified by the sampler network, the measured gene expression $\mathbf{Y}_\mathcal{V}$ is projected into gene features $\mathbf{H} \in \mathbb{R}^{|\mathcal{V}| \times d}$ via an MLP-based gene encoder.
These features, alongside the image features $\mathbf{X}$, are processed through independent network streams. 
For computational efficiency, both streams share the transformer parameters $\theta_\mathrm{enc}$ to align disparate modalities into a shared latent space, while utilizing independent layer normalization to accommodate their distinct distributions:
\begin{equation}
    \mathbf{X}_{\varepsilon} = f_{\mathrm{img}}(\mathbf{X}; \theta_\mathrm{enc}), \quad \mathbf{H}_{\varepsilon} = f_{\mathrm{gene}}(\mathbf{H}; \theta_\mathrm{enc}).
\end{equation}
The fused sequence $\mathbf{Z}_{\mathrm{fused}}$ is then constructed via a linear layer $\mathbf{W}_{\mathrm{fuse}} \in \mathbb{R}^{2d \times d}$. 
For visible spots ($i \in \mathcal{V}$), the concatenated gene and image features are projected into a joint representation through $\mathbf{W}_{\mathrm{fuse}}$. 
Similarly, for masked spots ($i \in \mathcal{M}$), a learnable mask token $\mathbf{m}_{\mathrm{token}}$ and the corresponding image features are integrated to guide the prediction process:
\begin{equation}
    \mathbf{z}_\mathrm{fused}^{(i)} = \begin{cases} \mathbf{W}_\mathrm{fuse}(\mathbf{H}_{\varepsilon, i} \oplus \mathbf{X}_{\varepsilon, i}) & \text{if } i \in \mathcal{V} \\ \mathbf{W}_\mathrm{fuse}(\mathbf{m}_\mathrm{token} \oplus \mathbf{X}_{\varepsilon, i}) & \text{if } i \in \mathcal{M} \end{cases}.
\end{equation}
Finally, $\mathbf{Z}_\mathrm{fused}$ is fed into the transformer decoder and an MLP head to predict the gene expression profiles $\hat{\mathbf{Y}}_\mathcal{M}$ for the masked set:
\begin{equation}
    \mathbf{Z}_\mathrm{dec} = f_{\mathrm{dec}}(\mathbf{Z}_\mathrm{fused}; \theta_\mathrm{dec}), 
    \quad
    \hat{\mathbf{y}}_i = \mathrm{MLP}(\mathbf{z}_\mathrm{dec}^{(i)}) \;\; \text{for } i \in \mathcal{M}.
\end{equation}

\myparagraph{Contrastive Feature Alignment.} Standard contrastive objectives based on binary supervision often introduce false conflicts by treating biologically similar semantic neighbors as negatives. 
Since our sampler selects contiguous regions, adjacent spots within $\mathcal{V}$ are highly likely to share similar morphological and transcriptomic profiles, which significantly exacerbates the risk of these false conflicts.
To mitigate this risk, we adopt the soft-target contrastive loss from $\mathrm{BLEEP}$~\cite{bleep}.
We first project the encoded features $\mathbf{X}_{\varepsilon}$ and $\mathbf{H}_{\varepsilon}$ for the visible spots ($i \in \mathcal{V}$) into $\tilde{\mathbf{X}}, \tilde{\mathbf{H}} \in \mathbb{R}^{|\mathcal{V}| \times c}$ via an MLP head.
Then, the soft target $\boldsymbol{\mathcal{T}}$ and the contrastive objective $\mathcal{L}_{\mathrm{con}}$ are defined as: 
\begin{equation} 
    \boldsymbol{\mathcal{T}} = \mathrm{softmax}\left( \frac{\tilde{\mathbf{X}}\tilde{\mathbf{X}}^T + \tilde{\mathbf{H}}\tilde{\mathbf{H}}^T}{2\tau} \right), \;
    \mathcal{L}_{\mathrm{con}} = \sum_{i \in \mathcal{V}} \frac{\left[ \mathrm{CE}(\boldsymbol{\mathcal{T}}_{i}, {\mathbf{p}}_{i}) + \mathrm{CE}(\boldsymbol{\mathcal{T}}_{i}, {\mathbf{q}}_{i}) \right]}{2|\mathcal{V}|},
\end{equation}
\noindent where $\mathbf{p}_{i} = \tilde{\mathbf{X}}_{i} \tilde{\mathbf{H}}^{T} / \tau$ and $\mathbf{q}_{i} = \tilde{\mathbf{H}}_{i} \tilde{\mathbf{X}}^T / \tau$ represent the cross-modal similarity logits for the $i$-th spot, and $\mathrm{CE}$ denotes the cross-entropy function.
This soft-alignment prevents biologically similar spots from being incorrectly penalized as negatives, yielding more biologically consistent joint representations.

\myparagraph{Objective Functions.} The overall training objective of \modelname{} is formulated as:
\begin{equation}
    \mathcal{L}_{\mathrm{total}} = \mathcal{L}_{\mathrm{pred}} + \lambda_{\mathrm{rank}}\mathcal{L}_{\mathrm{rank}} + \lambda_{\mathrm{con}}\mathcal{L}_{\mathrm{con}},
\end{equation}
where $\lambda_{\mathrm{rank}}$ and $\lambda_{\mathrm{con}}$ are hyperparameters balancing the contribution of each loss. 
The prediction loss $\mathcal{L}_{\mathrm{pred}}$ is defined as $\mathcal{L}_{\mathrm{mse}} + \lambda_{\mathrm{pcc}}\mathcal{L}_{\mathrm{pcc}}$, which represents the sum of mean squared error ($\mathcal{L}_{\mathrm{mse}}$) and Pearson correlation coefficient loss ($\mathcal{L}_{\mathrm{pcc}}$) weighted by $\lambda_{\mathrm{pcc}}$, both computed based on the predicted gene expression profiles for the masked spots $\hat{\mathbf{Y}}_\mathcal{M}$. 
\begin{table}[t]
\centering
\caption{Performance comparison under prediction and imputation settings. Best and second-best (prediction only) results are \textbf{bolded} and \underline{underlined}, respectively. The proposed framework is highlighted in gray. $r_{\mathcal{V}}$: ratio of measured genetic anchors.}
\label{tab:main}

\fontsize{8}{10}\selectfont 
\setlength{\tabcolsep}{1.5pt}

\begin{tabular}{l c ccc ccc ccc}
\toprule
\multirow{2}{*}{{\textbf{Method}}} & \multirow{2}{*}{\textbf{$r_{\mathcal{V}}$}} & \multicolumn{3}{c}{{\textbf{ST-Net}}} & \multicolumn{3}{c}{{\textbf{Her2ST}}} & \multicolumn{3}{c}{{\textbf{cSCC}}} \\
\cmidrule(lr){3-5} \cmidrule(lr){6-8} \cmidrule(lr){9-11}
 & & MSE $\downarrow$ & MAE $\downarrow$ & PCC $\uparrow$ & MSE $\downarrow$ & MAE $\downarrow$ & PCC $\uparrow$ & MSE $\downarrow$ & MAE $\downarrow$ & PCC $\uparrow$ \\
\midrule
BLEEP~\cite{bleep} & - & 0.3756 & 0.4736 & 0.0784 & 0.7426 & 0.6591 & 0.2747 & 0.6079 & 0.6013 & 0.4176 \\
HisToGene~\cite{histogene} & - & 0.2954 & 0.4269 & 0.1150 & 0.7005 & 0.6628 & 0.4284 & {0.6561} & {0.6677} & 0.2627 \\
Hist2ST~\cite{hist2st} & - & 0.3811 & 0.4822 & 0.1525 & 0.7843 & 0.7286 & 0.2479 & 1.0190 & 0.7639 & 0.3003 \\
THItoGene~\cite{thitogene} & - & 0.2925 & 0.4111 & 0.3666 & 0.8436 & 0.7069 & 0.3445 & 0.6798 & 0.6442 & 0.3897 \\
TRIPLEX~\cite{triplex} & - & 0.1472 & 0.2943 & 0.2320 & 0.8982 & 0.6946 & 0.3927 & \underline{0.4891} &\underline{0.5356} & 0.5416 \\
MERGE~\cite{merge} & - & \underline{0.1347} & \underline{0.2834} & \underline{0.6795} & \underline{0.6422} & \underline{0.6255} & \underline{0.5037} & 0.5353 & 0.5838 & \underline{0.5512} \\
PH2ST~\cite{ph2st} & 0.0 & {0.1599} & {0.3063} & {0.6479} & {0.7520} & {0.6798} & {0.3701} & 0.6421 & 0.6563 & {0.3198} \\
\rowcolor{gray!15} CAMMST & 0.0 & \textbf{0.1254} & \textbf{0.2740} & \textbf{0.6954} & \textbf{0.5769} & \textbf{0.5865} & \textbf{0.5534} & \textbf{0.4654} & \textbf{0.5350} & \textbf{0.5622} \\
\midrule
PH2ST & 0.1 & {0.1441} & {0.2759} & {0.6960} & {0.6791} & {0.6133} & {0.4735} & {0.5772} & {0.5909} & {0.4330} \\
\rowcolor{gray!15} CAMMST & 0.1 & \textbf{0.1119} & \textbf{0.2457} & \textbf{0.7337} & \textbf{0.5098} & \textbf{0.5224} & \textbf{0.6236} & \textbf{0.4137} & \textbf{0.4763} & \textbf{0.6273} \\
\midrule
PH2ST & 0.3 & {0.1120} & {0.2147} & {0.7839} & {0.5247} & {0.4760} & {0.6375} & {0.4504} & {0.4602} & {0.5992} \\
\rowcolor{gray!15} CAMMST & 0.3 & \textbf{0.0865} & \textbf{0.1905} & \textbf{0.8010} & \textbf{0.3876} & \textbf{0.4007} & \textbf{0.7315} & \textbf{0.3202} & \textbf{0.3686} & \textbf{0.7262} \\
\bottomrule
\end{tabular}
\end{table}
\section{Experiments}
\subsection{Experimental Setup}
\myparagraph{Datasets.} Experiments are conducted on three datasets: ST-Net~\cite{stnet}, Her2ST~\cite{her2st}, and cSCC~\cite{skin}. 
To ensure a fair comparison, we strictly follow the experimental protocols established by MERGE~\cite{merge}, utilizing 8-fold cross-validation with $\mathrm{SPCS}$ smoothing~\cite{spcs}, and targeting the top-250 highly expressed genes for prediction.

\myparagraph{Baselines and Evaluation Metrics.} We benchmark \modelname{} against several state-of-the-art baselines, including image-only prediction models and the spatial imputation framework, PH2ST~\cite{ph2st}. 
For a fair comparison, PH2ST is implemented using the same UNI2-h backbone as our framework. Performance is evaluated using three standard metrics: Pearson Correlation Coefficient (PCC), Mean Squared Error (MSE), and Mean Absolute Error (MAE).

\myparagraph{Evaluation Settings.} Performance is evaluated under two scenarios representing different levels of data availability: prediction and imputation.  
While the prediction setting relies solely on WSIs, the imputation setting incorporates a fraction of gene expression data as spatial anchors.
For the latter, predicted values in measured regions are replaced with ground truth to calculate slide-wide metrics, reflecting the overall prediction performance across the entire tissue.

\myparagraph{Implementation Details.} Our framework is trained using AdamW optimizer ($\mathrm{lr}=2e-4$, weight decay=$5e-2$).
During training, $10\%$ of the total spots are selected as visible spots by identifying $K=3$ regions, with each region containing an approximately equal number of spots.
Hyperparameters are assigned as $\beta=1.5$, $\lambda_{\mathrm{rank}}=0.001$, $\lambda_{\mathrm{con}}=0.05$, $\lambda_{\mathrm{pcc}}=0.5$, and $\tau=1.0$. 
Experiments are conducted on a single NVIDIA RTX A6000 GPU using the PyTorch framework.

\begin{figure}
\includegraphics[width=\textwidth]{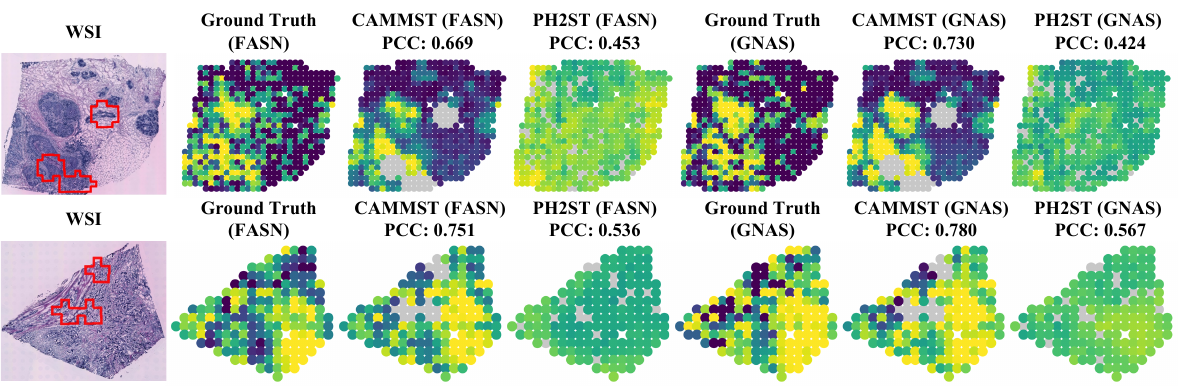}
\caption{Qualitative evaluation of gene expression prediction and sampling strategy. Red boundaries on the WSI indicate the contiguous regions selected by our sampler network. Spots in gray denote the corresponding visible spots.}
\label{fig:qualitative}
\end{figure}

\subsection{Main Results}

\myparagraph{Quantitative Results.} Table~\ref{tab:main} presents the comparative performance of our framework against state-of-the-art methods in both prediction and imputation settings. 
In the imputation setting, our framework outperforms the performance of PH2ST; for instance, on the cSCC dataset at $r_{\mathcal{V}}=0.1$, \modelname{} achieves a PCC of 0.6273 compared to 0.4330 for the baseline. 
These gains indicate that our framework effectively leverages contiguous genetic anchors to predict precise transcriptomic profiles. 
Notably, even in the prediction setting, \modelname{} consistently surpasses all image-only baselines, establishing a robust morphology-to-gene mapping even without the aid of genetic anchors.

\newcolumntype{Y}{>{\centering\arraybackslash}X}

\begin{table}[t!]
\centering
\caption{Comparison of computational efficiency on the Her2ST dataset. Both models utilize pre-computed histological features extracted from the UNI2-h backbone.}
\label{tab:efficiency}
\fontsize{8}{10}\selectfont 
\begin{tabularx}{\textwidth}{lYY}
\toprule
\textbf{Metric} & PH2ST & \modelname{} (Ours) \\
\midrule
Model Parameters (M)  & 290.73 & {44.59} \\
GFLOPs (per slide)  & 2386.21 $\pm$ 996.39 & {33.45 $\pm$ 13.94} \\
Training Time (per fold, hours)  & 7.91 $\pm$ 0.80 & {0.19 $\pm$ 0.03} \\
Training GPU Memory (GB)  & 10.54 $\pm$ 0.59 & {1.04 $\pm$ 0.01} \\
Inference Time (per slide, sec)  & 0.946 $\pm$ 0.443 & {0.021 $\pm$ 0.001} \\
Inference GPU Memory (GB)  & 5.93 $\pm$ 2.14 & {0.21 $\pm$ 0.02} \\
\bottomrule
\end{tabularx}
\end{table}

\myparagraph{Qualitative Results.} Fig.~\ref{fig:qualitative} presents a qualitative comparison of expression predictions for the tumor biomarker FASN~\cite{fasn} and the breast cancer biomarker GNAS~\cite{gnas}. 
While PH2ST produces over-smoothed patterns that fail to capture local variations, \modelname{} successfully recovers fine-grained gene expression that closely resembles the ground truth.
The WSI overlay illustrates how our sampler targets contiguous regions (red boundaries). 
By utilizing these visible spots (gray spots), our framework effectively guides the prediction process.

\myparagraph{Computational Efficiency.} Table~\ref{tab:efficiency} presents a comparison of computational efficiency. 
Compared to the baseline, our framework substantially reduces resource consumption across both training and inference. 
This advantage stems from the MAE-based architecture, which provides an optimal framework for the spatial imputation task.  
Consequently, \modelname{} achieves superior prediction performance and operates effectively without high-end hardware, providing a practical and high-precision solution for resource-constrained clinical environments.

\newcolumntype{Y}{>{\centering\arraybackslash}X}

\begin{table}[t]
\centering
\caption{Ablation study of each component on the Her2ST dataset.}
\label{tab:ablation}
\fontsize{8}{10}\selectfont 
\setlength{\aboverulesep}{0pt}
\setlength{\belowrulesep}{0pt}
\setlength{\tabcolsep}{8pt} 
\renewcommand{\arraystretch}{1.0}
\setlength{\extrarowheight}{2.5pt} 

\begin{tabularx}{\textwidth}{l Y Y Y} 
\toprule
\textbf{Method} & MSE $\downarrow$ & MAE $\downarrow$ & PCC $\uparrow$ \\
\midrule

\multicolumn{4}{>{\columncolor{gray!10}}l}{\textbf{\textit{Bio-Saliency Region Sampling}}} \\
\quad w/ Random Sampling (w/o sampler network) & 0.5205 & 0.5261 & 0.6106 \\
\quad w/ Regression Loss (w/o $\mathcal{L}_{\mathrm{rank}}$) & 0.5117 & 0.5225 & 0.6197 \\
\quad w/ Scattered Sampling (w/o region constraints) & 0.5204 & 0.5251 & 0.6119 \\

\specialrule{0pt}{0.5pt}{0.5pt} 

\multicolumn{4}{>{\columncolor{gray!10}}l}{\textbf{\textit{Contrastive Cross-Modal Fusion}}} \\
\quad w/o Contrastive Learning (w/o $\mathcal{L}_{\mathrm{con}}$) & 0.5150 & 0.5246 & 0.6171 \\
\quad w/ Standard Contrastive Learning (w/o soft-target) & 0.5118 & 0.5223 & 0.6212 \\

\midrule
Ours & 0.5098 & 0.5224 & 0.6236 \\
\bottomrule
\end{tabularx}
\end{table}

\subsection{Ablation Study} Table~\ref{tab:ablation} presents the ablation study results evaluating the impact of each module.
The sampler network, optimized with the proposed scale-aware ranking loss ($\mathcal{L}_{\mathrm{rank}}$), yields superior performance compared to random selection or regression-based $S_{\text{bio}}$ prediction. 
This improvement demonstrates that the ranking-based strategy identifies more effective genetic anchors while ensuring robustness against batch effects. 
A scattered sampling variant optimized identically but without regional constraints results in a performance decline, demonstrating that identifying informative points alone is insufficient.
This degradation suggests that the value of informational diversity is inherently tied to the surrounding spatial context. 
By selecting contiguous regions, our framework captures the essential signals for accurate transcriptomic prediction.
For cross-modal alignment, the soft-target contrastive loss outperforms standard or absent objectives. 
These results demonstrate that our approach effectively integrates multi-modal features by mitigating false conflicts between biologically related spots.
\section{Conclusion}
In this study, we proposed \modelname{}, a novel framework for gene expression prediction. 
Built upon the MAE paradigm, our model adaptively selects the most informative regions within the tissue and leverages measured gene expression to accurately predict whole-slide transcriptomic profiles. 
Beyond predictive performance, our framework is explicitly designed for practical efficiency, accommodating the physical constraints of real-world ST hardware and limited computing resources. 
Experimental results demonstrate that \modelname{} consistently outperforms all existing baselines, providing a highly efficient and accurate solution that establishes a practical foundation for routine clinical workflows.

\begin{credits}
\subsubsection{\ackname} This work was supported in part by the IITP RS-2024-00457882 (AI Research Hub Project), IITP 2020-II201361, NRF RS-2024-00345806, NRF RS-2023-002620, and RQT-25-120390.

\end{credits}
%
%
%
\bibliographystyle{splncs04}
\bibliography{reference}
\end{document}